\def\year{2017}\relax
\documentclass[letterpaper]{article}
\usepackage{aaai17,amsfonts,amsmath,times,helvet,courier,graphicx,float}
\newcommand\addtag{\refstepcounter{equation}\tag{\theequation}}
\newcommand{\ar}{\rightarrow}

\renewcommand{\c}{,\!}

\newcommand{\cF}{\mathcal{F}}
\newcommand{\cQ}{\mathcal{Q}}
\newcommand{\dconv}{\stackrel{d}{\rightarrow}}
\newcommand{\hA}{\hat{A}}
\newcommand{\hth}{\hat{\theta}}
\newcommand{\hW}{\hat{W}}
\newcommand{\iid}{\stackrel{iid}{\sim}}
\newcommand{\ind}{\stackrel{ind}{\sim}}
\newcommand{\la}{\lambda}
\newcommand{\pa}{\partial}
\newcommand{\Pconv}{\stackrel{P}{\rightarrow}}
\newcommand{\si}{\sigma}
\newcommand{\tth}{\tilde{\theta}}
\newcommand{\Th}{\Theta}
\renewcommand{\th}{\theta}
\newcommand{\vth}{\vartheta}

\title{Regularization for Unsupervised Deep Neural Nets}
\author{Baiyang Wang, Diego Klabjan\\
	Department of Industrial Engineering and Management Sciences, \\
	Northwestern University, 2145 Sheridan Road, C210\\
	Evanston, Illinois 60208\\
}

\begin{document}
\maketitle
\begin{abstract}
Unsupervised neural networks, such as restricted Boltzmann machines (RBMs) and deep belief networks (DBNs), are powerful tools for feature selection and pattern recognition tasks. We demonstrate that overfitting occurs in such models just as in deep feedforward neural networks, and discuss possible regularization methods to reduce overfitting. We also propose a ``partial'' approach to improve the efficiency of Dropout/DropConnect in this scenario, and discuss the theoretical justification of these methods from model convergence and likelihood bounds. Finally, we compare the performance of these methods based on their likelihood and classification error rates for various pattern recognition data sets.
\end{abstract}

\section{Introduction}
Unsupervised neural networks assume unlabeled data to be generated from a neural network structure, and have been applied extensively to pattern analysis and recognition. The most basic one is the restricted Boltzmann machine (RBM) \cite{SM07}, an energy-based model with a layer of hidden nodes and a layer of visible nodes. With such a basic structure, we can stack multiple layers of RBMs to create an unsupervised deep neural network structure, such as the deep belief network (DBN) and the deep Boltzmann machine (DBM) \cite{HO06,SH09}. These models can be calibrated with a combination of the stochastic gradient descent and the contrastive divergence (CD) algorithm or the PCD algorithm \cite{SM07,T08}. Once we learn the parameters of a model, we can retrieve the values of the hidden nodes from the visible nodes, thus applying unsupervised neural networks for feature selection. Alternatively, we may consider applying the parameters obtained from an unsupervised deep neural network to initialize a deep feedforward neural network (FFNN), thus improving supervised learning.

One essential question for such models is to adjust for the high-dimensionality of their parameters and avoid overfitting. In FFNNs, the simplest regularization is arguably the early stopping method, which stops the gradient descent algorithm before the validation error rate goes up. The weight decay method, or $L^s$ regularization, is also commonly used \cite{WF11}. Recently Dropout is proposed, which optimizes the parameters over an average of exponentially many models with a subset of all nodes \cite{SH14}. It has been shown to outperform weight decay regularization in many situations.

For regularizing unsupervised neural networks, sparse-RBM-type models encourage a smaller proportion of $1$-valued hidden nodes \cite{CI12,LE07}. DBNs are regularized in \citeauthor{GT13} (2013) with outcome labels. While these works tend to be goal-specific, we consider regularization for unsupervised neural networks in a more general setting. Our work and contributions are as follows: (1) we extend common regularization methods to unsupervised deep neural networks, and explain their underlying mechanisms; (2) we propose partial Dropout/DropConnect which can improve the performance of Dropout/DropConnect; (3) we compare the performance of different regularization methods on real data sets, thus providing suggestions on regularizing unsupervised neural networks. We note that this is the very first study illustrating the mechanisms of various regularization methods for unsupervised neural nets with model convergence and likelihood bounds, including the effective newly proposed partial Dropout/DropConnect.

Section 2 reviews recent works for regularizing neural networks, and Section 3 exhibits RBM regularization as a basis for regularizing deeper networks. Section 4 discusses the model convergence of each regularization method. Section 5 extends regularization to unsupervised deep neural nets. Section 6 presents a numerical comparison of different regularization methods on RBM, DBN, DBM, RSM \cite{SH09b} and Gaussian RBM \cite{SM07}. Section 7 discusses potential future research and concludes the paper.

\section{Related Works}
To begin with, we consider a simple FFNN with a single layer of input $\imath=(\imath_1,\ldots,\imath_I)^T$ and a single layer of output $o=(o_1,\ldots,o_J)^T\in\{0,1\}^J$. The weight matrix $W$ is of size $J\times I$. We assume the relation
\begin{equation}
E(o) = a(W\cdot \imath),
\end{equation}
where $a(\cdot)$ is the activation function, such as the sigmoid function $\si(x)=1/(1+e^{-x})$ applied element-wise. Equation (1) has the modified form in \citeauthor{SH14} (2014),
\[
\begin{cases}
E(o|m) = a(m\star (W\cdot \imath)),\\
m=(m_1,\ldots,m_J)^T\iid Ber(p), \addtag
\end{cases}
\]
where $\star$ denotes element-wise multiplication, and $Ber(\cdot)$ denotes the Bernoulli distribution, thereby achieving the Dropout (DO) regularization for neural networks. In Dropout, we minimize the objective function
\begin{equation}
-l_{DO}(W) = -\sum_{n=1}^NE_m[\log p(o^{(n)}|\imath^{(n)}, W, m)],
\end{equation}
which can be achieved by a stochastic gradient descent algorithm, sampling a different mask $m$ per data example $(o^{(n)},\imath^{(n)})$ and per iteration. We observe that this can be readily extended to deep FFNNs. Dropout regularizes neural networks because it incorporates prediction based on any subset of all the nodes, therefore penalizing the likelihood. A theoretical explanation is provided in \citeauthor{WW13} (2013) for Dropout, noting that it can be viewed as feature noising for GLMs, and we have the relation
\begin{equation}
-l_{DO}(W) \doteq -\sum_{n=1}^N\log p(o^{(n)}|\imath^{(n)}, W) + R^q(W).
\end{equation}
Here $J=1$ for simplicity, and $R^q(W)= \frac{1}{2}\frac{p}{1-p} \sum_{n=1}^N \linebreak \sum_{i=1}^IA''(W\imath^{(n)})(\imath_i^{(n)})^2 W_i^2$, where $A(\cdot)$ is the log-partition function of a GLM. Therefore, Dropout can be viewed approximately as the adaptive $L^2$ regularization \cite{BS13,WW13}. A recursive approximation of Dropout is provided in \citeauthor{BS13} (2013) using normalized weighted geometric means to study its averaging properties.

An intuitive extension of Dropout is DropConnect (DC) \cite{WZ13}, which has the form below
\[
\begin{cases}
E(o|m) = a((m\star W)\cdot \imath),\\
 m=(m_{ij})_{J\times I}\iid Ber(p), \addtag
\end{cases}
\]
and thus masks the weights rather than the nodes. The objective $l_{DC}(W)$ has the same form as in (3). There are a number of related model averaging regularization methods, each of which averages over subsets of the original model. For instance, Standout varies Dropout probabilities for different nodes which constitute a binary belief network \cite{BF13}. Shakeout adds additional noise to Dropout so that it approximates elastic-net regularization \cite{KL16}. Fast Dropout accelerates Dropout with Gaussian approximation \cite{WM13}. Variational Dropout applies variational Bayes to infer the Dropout function \cite{KS15}.

We note that while Dropout has been discussed for RBMs \cite{SH14}, to the best of our knowledge, there is no literature extending common regularization methods to RBMs and unsupervised deep neural networks; for instance, adaptive $L^s$ regularization and DropConnect as mentioned. Therefore, below we discuss their implementations and examine their empirical performance. In addition to studying model convergence and likelihood bounds, we propose partial Dropout/DropConnect which iteratively drops a subset of nodes or edges based on a given calibrated model, therefore improving robustness in many situations.

\section{RBM Regularization}
For a Restricted Boltzmann machine, we assume that $v=(v_1, \cdots, v_J)^T\in\{0,1\}^J$ denotes the visible vector, and $h=(h_1, \cdots, h_I)^T\in\{0,1\}^I$ denotes the hidden vector. Each $v_j$, $j=1,\ldots,J$ is a visible node
and each $h_i$, $i=1,\ldots,I$ is a hidden node. The joint probability is
\[
\begin{cases}
P(v,h) = e^{-E(v,h)}/\sum_{\nu,\eta}e^{-E(\nu,\eta)},\\
E(v,h) = -b^Tv-c^Th-h^TWv. \addtag
\end{cases}
\]
We let the parameters $\vth=(b, c, W)\in \Th$, which is a vector containing all components of $b$, $c$, and $W$. To calibrate the model is to find $\hth = \arg\max\limits_{\vth\in\Th}\sum_{n=1}^N \log P(v^{(n)}|\vth)$.

An RBM is a neural network because we have the following conditional probabilities
\[
\begin{cases}
P(h_i=1|v)=\si(c_i+W_{i\cdot}v),\\
P(v_j=1|h)=\si(b_j+W_{\cdot j}^Th), \addtag
\end{cases}
\]
where $W_{i\cdot}$ and $W_{\cdot j}$ represent, respectively, the $i$-th row and $j$-th column of $W$. The gradient descent algorithm is applied to calibration. The gradient of the log-likelihood can be expressed in the following form
\begin{equation}
-\frac{\pa \log P(v^{(n)})}{\pa \vth}=\frac{\pa \cF(v^{(n)})}{\pa \vth} - \sum_{v\in\{0,1\}^J}P(v)\frac{\pa \cF(v)}{\pa \vth},
\end{equation}
where $\cF(v)=-b^Tv-\sum_{i=1}^I\log(1+e^{c_i+W_{i\cdot}v})$ is the free energy. The right-hand side of (8) is approximated by contrastive divergence with $k$ steps of Gibbs sampling (CD-$k$) \cite{SM07}.

\subsection{Weight Decay Regularization}
Weight decay, or $L^s$ regularization, adds the term $\la\|W\|_s^s$ to the negative log-likelihood of an RBM. The most commonly used is $L^2$ (ridge regression), or $L^1$ (LASSO). In all situations, we do not regularize biases for simplicity. 

Here we consider a more general form. Suppose we have a trained set of weights $W$ from CD with no regularization. Instead of adding the term $\la\|W\|_s^s$, we add the term $\frac{\mu}{IJ}\sum_{i,j}|W_{ij}|^s/|\hW_{ij}|^s$ to the negative log-likelihood. Apparently this adjusts for the different scales of the components of $W$. We refer to this approach as adaptive $L^s$. We note that adaptive $L^1$ is the adaptive LASSO \cite{Z06}, and adaptive $L^2$ plus $L^1$ is the elastic-net \cite{Z05}. We consider the performance of $L^2$  regularization plus adaptive $L^1$ regularization ($L^2+AL^1$) below. 

\subsection{Model Averaging Regularization}
As discussed in \citeauthor{SH14} (2014), to characterize a Dropout (DO) RBM, we simply need to apply the following conditional distributions
\[
\begin{cases}
P_{DO}(h_i=1|v,m)=m_i\cdot\si(c_i+W_{i\cdot}v),\\ P_{DO}(v_j=1|h,m)=\si(b_j+W_{\cdot j}^Th). \addtag
\end{cases}                                            
\]             
Therefore, given a fixed mask $m\in\{0,1\}^I$, we actually obtain an RBM with all visible nodes $v$ and hidden nodes $\{h_i: m_i=1\}$. Hidden nodes $\{h_i: m_i=0\}$ are fixed to zero so they have no influence on the conditional RBM. Apart from replacing (7) with (9), the only other change needed is to replace $\cF(v)$ with $\cF_{DO}(v|m) = -b^Tv-\sum_{i=1}^Im_i\log(1+e^{c_i+W_{i\cdot}v})$. In terms of training, we suggest sampling a different mask per data example $v^{(n)}$ and per iteration as in \citeauthor{SH14} (2014). 

A DropConnect (DC) RBM is closely related; given a mask $m=\{0,1\}^{IJ}$ on weights $W$, $W$ in a plain RBM is replaced by $m\ast W$ everywhere. We suggest sampling a different mask $m$ per mini-batch since it is usually much larger than a mask in a Dropout RBM. 
\subsection{Network Pruning Regularization}
There are typically many nodes or weights which are of little importance in a neural network. In network pruning, such unimportant nodes or weights are discarded, and the neural network is retrained. This process can be conducted iteratively \cite{R93}. Now we consider two variants of network pruning for RBMs. For an trained set of weights $\hW$ with no regularization, we consider implementing a fixed mask $m=(m_{ij})_{I\times J}$ where
\begin{equation}
m_{ij} = 1_{|\hW_{ij}|\ge Q},\ Q = \cQ_{100(1-p)\%}(|\hW|),
\end{equation}
i.e. $Q$ is the $100(1-p)\%$-th left percentile of all $|\hW_{ij}|$, and $p\in(0,1)$ is some fixed proportion of retained weights. We then recalibrate the weights and biases fixing mask $m$, leading to a simple network pruning (SNP) procedure which deletes $100(1-p)\%$ of all weights. We may also consider deleting $100(1-p)/r\%$ of all weights at a time, and conduct the above process $r$ times, leading to an iterative network pruning (INP) procedure. 

\subsection{Hybrid Regularization}
We may consider combining some of the above approaches. For instance, \citeauthor{SH14} (2014) considered a combination of $L^s$ and Dropout. We introduce two new hybrid approaches, namely partial DropConnect (PDC) presented in Algorithm 1 and partial Dropout (PDO), which generalizes DropConnect and Dropout, and borrows from network pruning. The rationale comes from some of the model convergence results exhibited later.

As before, suppose we have a trained set of weights $\hW$ with no regularization. Instead of implementing a fixed mask $m$, we perform DropConnect regularization with different retaining probabilities $p_{ij}$ for each weight $W_{ij}$. We let the quantile $Q = \cQ_{100(1-q)\%}(|\hW|)$, and
\begin{equation}
p_{ij} = 1_{|\hW_{ij}|\ge Q} + p_0 \cdot 1_{|\hW_{ij}|< Q}.
\end{equation}

Therefore, we sample a different $m=(m_{ij})_{I\times J}\ind Ber(p_{ij})$ per mini-batch, which means that we always keep $100q\%$ of all the weights, and randomly drop the remaining weights with probability $100(1-p_0)\%$. The mask $m$ can be resampled iteratively. Intuitively, we are trying to maximize the following
\begin{equation} 
\max_{\vth\in\Th, p_{ij}\in\{p_0,1\}} E_m [\log P(data|\vth, m)].
\end{equation}
such that $m_{ij}\ind Ber(p_{ij})$, and $\sum{1_{p_{ij}=1}} = qIJ$.
\\

\fbox{
\begin{minipage}[c][11em][t]{0.42\textwidth}
{\bf Algorithm 1}. (Partial DropConnect)
\begin{enumerate}
\item Initialize $\hth_p = \hth$, the unregularized trained parameters for an RBM.
\item Find retaining rates $p = (p_{ij})_{I\times J}$ from (11).
\item Retrain weights $\vth$ with DropConnect for a given number of iterations, and then update $\hth_p$.
\item If maximum number of iterations reached, stop and obtain $\hth_p$; otherwise, go back to Step 2. 
\end{enumerate}
\end{minipage}
}
\\

This technique is proposed because we hypothesize that some weights could be more important than others {\it a posteriori}, so dropping them could cause much variation among the models being averaged. From (11), in partial Dropout, we tend to drop weights which have smaller magnitude, since setting larger weights to zero may substantially alter the structure of a neural network. Experiments on real data show that this technique can effectively improve the performance of plain DropConnect.

We denote $l_p(\vth) = E_{m\sim Ber(p)} [\log P(data|\vth, m)]$, and $l(\vth) = \log P(data|\vth)$. From first-order Taylor's expansion, 
\begin{align*}
&\hspace{1em} |l_p(\hth_p)-l(\hth)| \le |l_p(\hth_p)-l(\hth_p)| + |l(\hth_p)-l(\hth)|\\
&= \left|\sum_{i,j}\frac{\pa}{\pa W_{ij}}l(\tth)(1-p_{ij})\hW_{ij,p}\right|  + |l(\hth_p)-l(\hth)| \\
&\le K\sum(1-p_{ij})|\hW_{ij,p}| + |l(\hth_p)-l(\hth)|. \addtag   
\end{align*} 
Here $\tth$ lies between $\hth_p$ and $m\ast\hth_p$ from Taylor's expansion, and $K=\sup_{\vth\in\Th} \|\frac{\pa}{\pa\vth}l(\vth)\|_\infty$ is a Lipschitz constant. 
 
Note that given $p$ and $\hth_p$, Step 2 in Algorithm 1 lowers the term $\sum(1-p_{ij})|\hW_{ij,p}|$ by assigning $(1-p_0)$ to weights of smaller magnitude, reducing an upper bound of $|l_p(\hth_p)-l(\hth)|$. Step 3 further increases $l_p(\hth_p)$ and reduces the gap $|l_p(\hth_p)-l(\hth)|$. Therefore, each iteration of Algorithm 1 tends to increase $l_p(\hth_p)$, and hence Algorithm 1 provides an intuitive solution to problem (12). 

We also consider a partial Dropout approach which is analogous to partial DropConnect and keeps some important nodes rather than weights. We set a mask for nodes $m=(m_1, \ldots, m_I)$, $m_i\ind Ber(p_i)$, where
\[
\begin{cases}
p_i = 1_{\|\hW_{i\cdot}\|\ge Q} + p_0 \cdot 1_{\|\hW_{i\cdot}\|< Q}, \\
Q = \cQ_{100(1-q)\%}(\|\hW_{i\cdot}\|). 
\end{cases} \addtag
\]
This algorithm protects more important hidden nodes from being dropped in order to reduce variation. We also evaluate its empirical performance later.

\section{More Theoretical Considerations}
Here we discuss the model convergence properties of different regularization methods when the number of data examples $N\ar\infty$. We mark all regularization coefficients and parameter estimates with $ ^{(N)}$ when there are $N$ data examples. We assume $\vth=(b,c,W)\in \Th$, which is compact, $\dim(\Th)=D$, $P(v|\vth)$ is unique for each $\vth\in\Th$, and $v^{(1)},\ldots,v^{(N)}$ are i.i.d. generated from an RBM with a ``true'' set of parameters $\th$. We denote each regularized calibrated set of parameters as $\tth^{(N)}$.

Let $A = \{d: \th_d\ne0\}$ and $\th_A = \{\th_d: d\in A\}$. \cite{Z06} showed that $AL^1$ guarantees asymptotic normality and identification of set $A$ for linear regression. We demonstrate that similar results hold for $L^2+AL^1$ for RBMs. We let $\la^{(N)}=(\la_1^{(N)},\ldots,\la_D^{(N)})$ and  $\mu^{(N)}=(\mu_1^{(N)},\ldots,\mu_D^{(N)})$ be the $L^2$ and $L^1$ regularization coefficients for each component. The proofs of all propositions and corollaries below are in the supplementary material \cite{WK16}.

{\it Proposition 1}. (a) If $\la^{(N)}/N\ar0$, $\mu^{(N)}/N\ar0$ as $N\ar\infty$, then the estimate $\tth^{(N)}\Pconv\th$; (b) if also, $\mu_d^{(N)}/\sqrt{N}\ar0\cdot1_{\th_d\ne0}+\infty\cdot1_{\th_d=0}$,  $\la^{(N)}/\sqrt{N}\ar0$, then $\sqrt{n}(\tth_A^{(N)}-\th_A)\dconv N(0,I^{-1}(\th_A))$, where $I$ is the Fisher information matrix; $P(\hA^{(N)} = A )\ar1$, where $\hA^{(N)} = \{d: \tth_d^{(N)}\ne0\}$. \hfill $\square$

For Dropout and DropConnect RBMs, we also assume that the data is generated from a plain RBM structure. We assume $p^{(N)}$ is of size $I\times J$ as in (11) for DropConnect and of length $I$ as in (14) for Dropout, therefore covering the cases of both original and partial Dropout/DropConnect with a fixed set of dropping rates. With a decreasing dropping rate $1-p^{(N)}\ar0$ with $N\ar\infty$, we obtain the following convergence result. 

{\it Proposition 2}. If $p^{(N)}\ar1$ as $N\ar\infty$, then $\tth^{(N)}\Pconv\th$.

\hfill $\square$

For network pruning, we show that as the number of data examples increase, if the retained proportion of parameters $p^{(N)} = p$ can cover all nonzero components of $\th$, we will not miss any important component.

{\it Proposition 3}. Assume $p>p_0:=|A|/D$. Then for simple network pruning, as $N\ar\infty$, (a) $\tth^{(N)}\Pconv\th$; (b) for sufficiently large $N$, there exists $\rho > 0$ such that $P(A\in\hA^{(N)}) \ge 1 - e^{-\rho N}$. \hfill $\square$

{\it Corollary 1}. The above results also hold for iterative network pruning. \hfill $\square$ 

We note that for all regularization methods, under the above conditions, the calibrated weights converge to the ``true'' set of parameters $\th$, which indicates consistency. Also, adding $L^1$ regularization guarantees that we can identify components of zero value with infinitely many examples. The major benefits of Dropout come from the facts that it makes $L^2$ regularization adaptive, and also encourages more confident prediction of the outcomes \cite{WW13}. We propose partial DropConnect also based on Proposition 3, i.e. we do not drop the more important components of $\th$, therefore possibly reducing variation caused by dropping influential weights. Partial Dropout follows from the same reasoning.

\section{Extension to Other Networks}
\subsection{Deep Belief Networks}
We consider the multilayer network below,
\begin{equation}
P(v,h^1,\ldots,h^L) \ P(v|h^1)\prod_{l=1}^{L-1}P(h^l|h^{l+1})P(h^L),
\end{equation}
where each probability on the right-hand side is from an RBM. To train the weights of $\textrm{RBM}(v,h_1)$, $\ldots$, $\textrm{RBM}(h^{L-1},h^L)$, we only need to carry out a greedy layer-wise training approach, i.e. we first train the weights of $\textrm{RBM}(v,h_1)$, and then use $E(h^1|v)$ to train $\textrm{RBM}(h_1, h_2)$, etc. The weights of the RBMs are used to initialize a deep FFNN which is finetuned with gradient descent. RBM regularization is applicable to each layer of a DBN.

Here we show that adding layers to a Dropout/Drop-Connect DBN improves the likelihood given symmetry of the weights of two adjacent layers. Similar results for plain DBN are in \citeauthor{HO06} (2006) and \citeauthor{B07} (2007). We demonstrate this by using likelihood bounds.

We let $DBN_L$ denote an $L$-layer DBN and $DBN_{L+1}$ denote an $(L+1)$-layer DBN with the first $L$ layers being the same as in $DBN_L$. For a data example of a visible vector $v$, the log-likelihood is bounded as follows,
\begin{align*}
&{\hspace{1.5em}} E_m[\log P_{DBN_{L+1}}(v|m,m^\ast)] \\
&\ge E_m[H_{P_{DBN_L}(h^L|v,m)}] +\sum_{h^L} E_{m,m^\ast}\{P_{DBN_L}(h^L|v,m)\\ 
&{\hspace{1em}} \cdot [\log P_{RBM_{L+1}}(h^L|m^\ast)+\log P_{DBN_L}(v|h^L,m)]\}. \addtag
\end{align*}

Here, $H$ is the entropy function, and the derivation is analogous to Section 11 in \citeauthor{B07} (2007). Mask $m$ is for $DBN_L$, and mask $m^\ast$ is for the new $(L+1)$-th layer. Note that after we have trained the first $L$ layers, and initialized the $(L+1)$-th layer symmetric to the $L$-th layer, assuming a constant dropping probability, we have
\begin{equation}
E_{m^\ast}[\log P_{RBM_{L+1}}(h^L|m^\ast)] = E_{m}[\log P_{DBN_L}(h^L|m)],
\end{equation}
so $DBN_{L+1}$ has the same log-likelihood bound as $DBN_L$. Training $RBM_{L+1}$, $E_{m^\ast}[\log P_{RBM_{L+1}}(h^L|m^\ast)]$ is guaranteed to increase, and therefore the likelihood of $DBN_{L+1}$ is expected to improve. As a result, for regularized unsupervised deep neural nets, adding layers also tend to elevate the explanatory power of the network. Adding nodes has the same effect, providing a rationale for deep and large-scale networks. We present the following proposition.

{\it Proposition 4}. Adding nodes or layers (preserving weight symmetry) to a Dropout/DropConnect DBN continually improves the likelihood; also, adding layers of size $J\le H^1\le H^2\le \cdots$ continually improves the likelihood. \hfill $\square$

\subsection{Other RBM Variants}
More descriptions of DBMs, RSMs, and Gaussian RBMs are in the supplementary material \cite{WK16}. RBM regularization can be extended to all these situations.

\section{Data Studies}
In this section, we compare the empirical performance of the aforementioned regularization methods on the following data sets: MNIST, NORB (image recognition); 20 Newsgroups, Reuters21578 (text classification); ISOLET (speech recognition). All results are obtained using GeForce GTX TITAN X in Theano.

\subsection{Experiment Settings}
We consider the following unsupervised neural network structures: DBN/DBM for MNIST; DBN for NORB; RSM plus logistic regression for 20 Newsgroups and Reuters21578; GRBM for ISOLET. CD-$1$ is performed for the rest of the paper. The following regularization methods are considered: None (no regularization); DO; DC; $L^2$; $L^2+AL^1$; SNP; INP($r=3$); PDO; PDC. The number of pretraining epochs is $100$ per layer and the number of finetuning epochs is $300$, with a finetuning learning rate of $0.1$. For $L^2+AL^1$, SNP, and INP which need re-calibration, we cut the $100$ epochs into two halves ($4$ quarters for INP). For regularization parameters, we apply the following ranges: $p=0.8\sim0.9$ for DO/DC/SNP/INP; $\la=10^{-5}\sim10^{-4}$ for $L^2$, similar to \citeauthor{H10} (2010); $\mu=0.01\sim0.1$ for $L^2+AL^1$; $p_0=0.5$, $q=0.7\sim0.9$ or the reverse for PDO/PDC. We only make one update to the ``partial'' dropping rates to maintain simplicity. From the results, we note that unsupervised neural networks tend to need less regularization than FFNNs. We choose the best iteration and regularization parameters over a fixed set of parameter values according to the validation error rates. 

\subsection{The MNIST Data Set}
The MNIST data set consists of $28^2$ pixels of handwritten $0$-$9$ digits. There are $50\c000$ training examples, $10\c000$ validation and $10\c000$ testing examples. We first consider the likelihood of the testing data of an RBM with $500$ nodes for MNIST. There are two model fitting evaluation criteria: pseudo-likelihood and AIS-likelihood \cite{SM08}. The former is a sum of conditional likelihoods, while the latter directly estimates $P(v)$ with AIS. 

In Figure 1 below using log-scale, $p=0.9$ for DO, and $\la=10^{-4}$ for $L^2$. These figures tend to be representative of the model fitting process. The pseudo-likelihood is a more optimistic estimate of the model fitting. We observe that Dropout outperforms the other two after about $50$ epochs, and $L^2$ regularization does not improve the pseudo-likelihood. In terms of the AIS-likelihood, which is a much more conservative estimate of the model fitting, the fitting process seems to have three stages: (1) initial fitting; (2) ``overfitting''; (3) re-fitting. We observe that $L^2$ improves the likelihood significantly, while Dropout catches up at about $300$ epochs. Therefore, Dropout tends to improve model fitting according to both likelihood criteria.
\begin{figure}[H]
	\centering
	\includegraphics[scale=1.2]{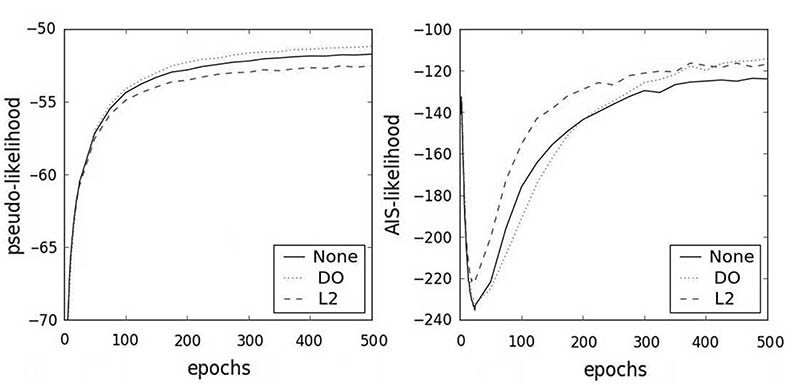}
	\caption{Left: Pseudo-likelihood of the RBM over $500$ pretraining epochs. Right: AIS-likelihood of the RBM over $500$ pretraining epochs. }
\end{figure}
In Figure 2, we can observe that more nodes increase the pseudo-likelihood, which is consistent with Proposition 4, but exhibit ``overfitting'' for the AIS-likelihood. However, such ``overfitting'' does not exist for pretraining purposes as well. Thus we suggest the pseudo-likelihood, and the AIS-likelihood should be viewed as too conservative.
\begin{figure}[H]
	\centering
	\includegraphics[scale=0.75]{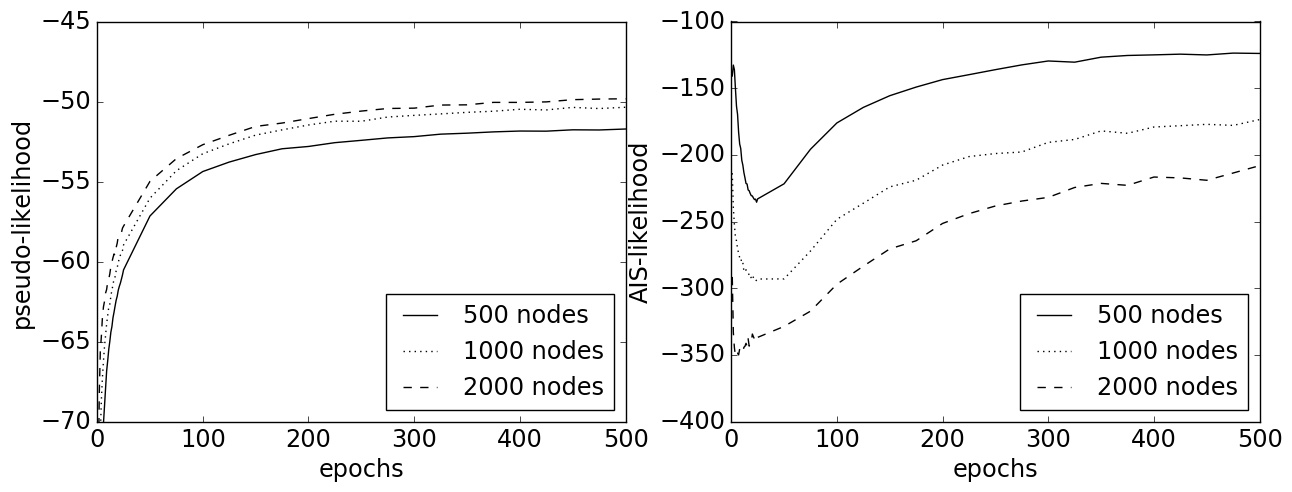}
	\caption{Left: Pseudo-likelihood of the Dropout RBM over $500$ pretraining epochs. Right: AIS-likelihood of the Dropout RBM over $500$ pretraining epochs. }
\end{figure}
Classification error rates tend to be a more practical measure. We first consider a $3$-hidden-layer DBN with $1\c000$ nodes per layer, pretraining learning rate $0.01$, and batch size $10$; see Table 1. We tried DBNs of $1$, $2$, and $4$ hidden layers and found the aforementioned structure to perform best with None as baseline. The same was done for all other structures. We calculate the means of the classification errors for each regularization method averaged over $5$ random replicates and their standard deviations. In each table, we stress in bold the top $3$ performers with ties broken by deviation. We note that most of the regularization methods tend to improve the classification error rates, with DC and PDO yielding slightly higher error rates than no regularization. 

\begin{table}[H]
\centering
\begin{tabular}{cccccccccc}
\hline & None & DO & DC & $L^2$ & $L^2+AL^1$ \\ 
\hline m. & $1.35\%$ & ${\bf 1.30\%}$ & $1.37\%$ & $1.31\%$ & $1.35\%$ \\  
\hline sd. & $0.02\%$ & $0.04\%$ & $0.02\%$ & $0.03\%$ & $0.01\%$ \\ 
\hline & SNP & INP & PDO & PDC \\ 
\hline m. & ${\bf 1.30\%}$ & $1.32\%$ & $1.36\%$ & ${\bf 1.30\%}$ \\  
\hline sd. & $0.03\%$ & $0.03\%$ & $0.04\%$ & $0.03\%$ \\ 
\hline
\end{tabular}
\caption{Classification errors for a $3$-layer DBN for the MNIST data set.}
\end{table}

In Table 2, we consider a $3$-hidden-layer DBM with $1\c000$ nodes per layer. For simplicity, we only classify based on the original features. We let the pretraining learning rate be $0.03$ and the batch size be $10$.

\begin{table}[H]
\centering
\begin{tabular}{cccccccccc}
\hline & None & DO & DC & $L^2$ & $L^2+AL^1$ \\  
\hline m. & $1.22\%$ & $1.21\%$ & $1.20\%$ & ${\bf 1.14\%}$ & ${\bf 1.15\%}$ \\  
\hline sd. & $0.02\%$ & $0.02\%$ & $0.02\%$ & $0.02\%$ & $0.04\%$ \\ 
\hline & SNP & INP & PDO & PDC \\ 
\hline m. & $1.18\%$ & $1.26\%$ & $1.21\%$ & ${\bf 1.12\%}$ \\  
\hline sd.  & $0.03\%$ & $0.02\%$ & $0.03\%$ & $0.02\%$ \\ 
\hline 
\end{tabular}
\caption{Classification errors for a $3$-layer DBM for the MNIST data set.}
\end{table}
It can be observed that regularization tends to yield more improvement for DBM than DBN, possibly because a DBM doubles both the visible layer and the third hidden layer, resulting in a ``larger'' neural network structure in general. Only INP proves to be unsuitable for the DBM; all other regularization methods work better, with PDC being the best.

\subsection{The NORB Data Set}
The NORB data set has $5$ categories of images of 3D objects. There are $24\c300$ training examples, with $2\c300$ validation examples held out, and $24\c300$ testing examples. We follow preprocessing of \citeauthor{NH09} (2009), and apply a sparse two-hidden-layer DBN with $4\c000$ nodes per layer as in \citeauthor{LE07} (2007) with a sparsity regularization coefficient of $10.0$ and the first hidden layer being a Gaussian RBM. The pretraining learning rates are $0.001$ and $0.01$ for the first and second hidden layer, and the batch sizes for pretraining and finetuning are $100$ and $20$. Because the validation error often goes to zero, we choose the $300$-th epoch and fix the regularization parameters as follows based on the best values of other data sets: $p=0.9$ for DO/DC/SNP/INP, $\la=10^{-4}$ for $L^2$, $\mu=0.1$ for $L^2+AL^1$, $(p_0, q)=(0.5, 0.8)$ for PDO and $(p_0,q)=(0.8,0.5)$ for PDC. In Table 3, only weight decay and PDO/PDC perform better than None, with PDC again being the best.

\begin{table}[H]
	\setlength{\tabcolsep}{0.4em}
	\centering
	\begin{tabular}{cccccccccc}
		\hline & None & DO & DC & $L^2$ & $L^2+AL^1$ \\ 
		\hline m. & $11.00\%$ & $11.15\%$ & $11.19\%$ & ${\bf 10.93\%}$ & ${\bf 10.91\%}$ \\  
		\hline sd. & $0.15\%$ & $0.12\%$ & $0.10\%$ & $0.18\%$ & $0.17\%$ \\ 
		\hline & SNP & INP & PDO & PDC & \\ 
		\hline m. & $11.04\%$ & $11.14\%$ & $10.95\%$ & ${\bf 10.81\%}$ & \\  
		\hline sd. & $0.18\%$ & $0.20\%$ & $0.15\%$ & $0.13\%$ & \\ 
		\hline 
	\end{tabular}
	\caption{Classification errors for the NORB data set.}
\end{table}

\subsection{The 20 Newsgroups Data Set}
The 20 Newsgroups data set is a collection of news documents with $20$ categories. There are $11\c293$ training examples, from which $6\c293$ validation examples are randomly held out, and $7\c528$ testing examples. We adopt the stemmed version, retain the most common $5\c000$ words, and train an RSM with $1\c000$ hidden nodes in a single layer. We consider this as a simple case of deep learning since it is a two-step procedure. The pretraining learning rate is $0.02$ and the batch size is $50$. We apply logistic regression to classify the trained features, i.e. hidden values of the RSM, as in \citeauthor{SS13} (2013). This setting is quite challenging for unsupervised neural networks. In Table 4, Dropout performs best with other regularization methods yielding improvements except DropConnect.
\begin{table}[H]
	\centering
	\begin{tabular}{cccccccccc}
		\hline  & None & DO & DC & $L^2$ & $L^2+AL^1$ \\ 
		\hline m. & $30.8\%$ & ${\bf 28.8\%}$ & $35.2\%$ & $30.1\%$ & $30.1\%$\\  
		\hline sd. & $0.70\%$ & $0.23\%$ & $0.91\%$ & $0.30\%$ & $0.65\%$\\ 
		\hline  & SNP & INP & PDO & PDC \\ 
		\hline m.  & ${\bf 29.7\%}$ & $29.7\%$ & $30.1\%$ & ${\bf 29.7\%}$ \\  
		\hline sd. & $0.26\%$ & $0.48\%$ & $0.71\%$ & $0.34\%$ \\ 
		\hline 
	\end{tabular}
	\caption{Classification errors for the trained features of RSM for the 20 Newsgroups data set.}
\end{table}
\subsection{The Reuters21578 Data Set}
The Reuters21578 data set is a collection of newswire articles. We adopt the stemmed R-52 version which has $52$ categories, $6\c532$ training examples, from which $1\c032$ validation examples are randomly held out, and $2\c568$ testing examples. We retain the most common $2\c000$ words, and train an RSM with $500$ hidden nodes in a single layer. The pretraining learning rate is $0.1$ and the batch size is $50$. We make the learning rate large because the cost function is quite bumpy. From Table 5, we note that PDC works best, and PDO improves the performance of Dropout.

\begin{table}[H]
	\setlength{\tabcolsep}{0.5em}
	\centering
	\begin{tabular}{cccccc}
		\hline & None & DO & DC & $L^2$ & $L^2+AL^1$  \\ 
		\hline m. & $10.50\%$ & $11.91\%$ & $10.10\%$ & $10.06\%$ & $9.99\%$\\
		\hline sd. & $0.64\%$ & $0.70\%$ & $0.32\%$ & $0.28\%$ & $0.41\%$\\
		\hline & SNP & INP & PDO & PDC \\ 
		\hline m. & ${\bf 9.99\%}$ & $10.10\%$ & ${\bf 9.98\%}$ & ${\bf 9.84\%}$ \\  
		\hline sd. & $0.27\%$ & $0.30\%$ & $0.24\%$ & $0.23\%$ \\ 
		\hline 
	\end{tabular}
	\caption{Classification errors for the trained features from RSM and the Reuters21578 data set.}
\end{table}
\subsection{The ISOLET Data Set}
The ISOLET data set consists of voice recordings of the Latin alphabet (a-z). There are $6\c138$ training examples, from which $638$ validation examples are randomly held out, and $1\c559$ testing examples. We train a $1\c000$-hidden-node Gaussian RBM with pretraining learning rate $0.005$, batch size $20$, and initialize a FFNN, which can be viewed as a single-hidden-layer DBN. From Table 6, it is evident that all regularization methods work better then None, with PDC again being the best.

\begin{table}[H]
	\centering
	\begin{tabular}{cccccccccc}
		\hline  & None & DO & DC & $L^2$ & $L^2+AL^1$  \\ 
		\hline m. & $3.98\%$ & $3.87\%$ & $3.88\%$ & ${\bf 3.83\%}$ & ${\bf 3.86\%}$ \\  
		\hline sd. & $0.09\%$ & $0.06\%$ & $0.11\%$ & $0.10\%$ & $0.07\%$  \\
		\hline  & SNP & INP & PDO & PDC \\ 
		\hline m. & ${\bf 3.86\%}$ & $3.86\%$ & $3.96\%$ & ${\bf 3.78\%}$ \\  
		\hline sd. & $0.07\%$ & $0.10\%$ & $0.08\%$ & $0.05\%$ \\ 
		\hline 
	\end{tabular}
	\caption{Classification errors for the ISOLET data set.}
\end{table}

\subsection{Summary}
From the above results, we observe that regularization does improve the structure of unsupervised deep neural networks and yields lower classification error rates for each data set studied herein. The most robust methods which yield improvements for all six instances are $L^2$, $L^2+AL^1$, and PDC. SNP is also acceptable, and preferable over INP. PDO can yield improvements for Dropout when Dropout is unsuitable for the network structure. PDC turns out to be the most stable method of all, and thus the recommended choice.

\section{Conclusion}
Regularization for deep learning has aroused much interest, and in this paper, we extend regularization to unsupervised deep learning, i.e. for DBNs and DBMs. We proposed several approaches, demonstrated their performance, and empirically compared the different techniques. For the future, we suggest that it would be of interest to consider more variants of model averaging regularization for supervised deep learning as well as novel methods of unsupervised learning; for instance, \citeauthor{KW14} (2015) provided an interesting variational Bayesian auto-encoder approach.

\bibliography{b1}
\bibliographystyle{aaai}
\end{document}